\documentclass[10pt]{article}
 
\usepackage[margin=1in]{geometry}
\usepackage[mathscr]{eucal}
\usepackage{amsmath,amsthm,amssymb}
\usepackage{authblk}
\usepackage{mathtools}
\usepackage{algorithm,algorithmic}
\usepackage{booktabs} 
\usepackage{hyperref}
\usepackage{url}
\usepackage{natbib}
\usepackage{footnote}
\usepackage{graphicx}
\usepackage{subfig}
\usepackage{bm}
\usepackage{breqn}

\usepackage{paralist}
\usepackage{calrsfs}
\usepackage{subfiles}
\usepackage{xcolor}
\theoremstyle{plain}
\newtheorem{theorem}{Theorem}
\newtheorem{corollary}{Corollary}[theorem]
\newtheorem{lemma}[theorem]{Lemma}

\newtheorem{definition}{Definition}
\newtheorem{remark}{Remark}

\newtheorem*{theorem*}{Theorem}

\newtheorem{innercustomgeneric}{\customgenericname}
\providecommand{\customgenericname}{}
\newcommand{\newcustomtheorem}[2]{%
  \newenvironment{#1}[1]
  {%
   \renewcommand\customgenericname{#2}%
   \renewcommand\theinnercustomgeneric{##1}%
   \innercustomgeneric
  }
  {\endinnercustomgeneric}
}

\newcustomtheorem{customthm}{Theorem}
\newcustomtheorem{customlemma}{Lemma}

\newcommand{\bb}[1]{\mathbb{#1}}

\newcommand{\wh}[1]{\widehat{#1}}
\newcommand{\bfa}[1]{\boldsymbol{#1}}
\newcommand{\fk}[1]{\mathfrak{#1}}
\DeclareMathAlphabet{\pazocal}{OMS}{zplm}{m}{n}
\newcommand{\ca}[1]{\pazocal{#1}}

\newcommand{\pa}[1]{\mathcal{#1}}

\newcommand{\tz}[1]{\fk S_{#1}}

\newcommand{\dtz}[2]{\fk S^{#1}_{#2}}
\newcommand{\ewt}[2]{\bb E[\Vert\delta^i\bfa w_{#1}\Vert^{#2}]}
\newcommand{\vwt}[1]{Var[\Vert\delta^i\bfa w_{#1}\Vert]}
\newcommand{\wt}[1]{\Vert\delta^i\bfa w_{#1}\Vert}
\newcommand{\tda}{\tilde{\bfa A}}

\newcommand{\emrr}[1]{\wh R_{#1}(h_{#1})}
\newcommand{\mrr}[1]{R(h_{#1})}

\newcommand{\st}[1]{{#1}^*}
\newcommand{\stp}[1]{{#1}^{*\prime}}
\newcommand{\stpb}[1]{{\bfa #1}^{*\prime}}
\newcommand{\tdas}{\tda^{*}}
\newcommand{\pr}[1]{{#1}^\prime}

\newcommand{\mr}[1]{ R(h_{\ca Z^{#1}})}
\newcommand{\bbem}{\bb E_{\ca Z_1^m\sim\bb D^m}}
\newcommand{\bbemt}{\bb E_{\ca Z_1^m,\ca Z^\text{test}\sim\bb D^{m+1}}}
\newcommand{\tet}{\text{test}}
\newcommand{\wha}{\wh{\bfa A}}

\newcommand{\dw}{\Vert\delta^i\bfa w_t\Vert}
\newcommand{\dwt}{\Vert\delta^i\bfa w_T\Vert}

\DeclareMathOperator*{\argmax}{arg\,max}

\begin{document}
\title{Multi-fidelity Stability for Graph Representation Learning}
\author[1]{Yihan He\thanks{yihan.he@nyu.edu}}
\author[1,2]{Joan Bruna\thanks{bruna@cims.nyu.edu}}
\affil[1]{Courant Institute, New York University}
\affil[2]{Center of Data Science, New York University}
\maketitle
\begin{abstract}
In the problem of structured prediction with graph representation learning (GRL for short), the hypothesis returned by the algorithm maps the set of features in the \emph{receptive field} of the targeted vertex to its label. To understand the learnability of those algorithms, we introduce a weaker form of uniform stability termed \emph{multi-fidelity stability} and give learning guarantees for weakly dependent graphs. We testify that ~\citet{london2016stability}'s claim on the generalization of a single sample holds for GRL when the receptive field is sparse. In addition, we study the stability induced bound for two popular algorithms: \textbf{(1)} Stochastic gradient descent under convex and non-convex landscape. In this example, we provide non-asymptotic bounds that highly depend on the sparsity of the receptive field constructed by the algorithm. \textbf{(2)} The constrained regression problem on a 1-layer linear equivariant GNN. In this example, we present lower bounds for the discrepancy between the two types of stability, which justified the multi-fidelity design.
\end{abstract}
\section{Introduction}
The problem of structured prediction of networks has been extensively studied due to the rich literature of social graphs and network structures in the real world. In this problem, the goal is to infer the label of its vertices, given some past observations. Recent progress in graph representation learning (GRL for short) achieved remarkable improvement over standard methods in this task. Typical applications of these algorithms include chemo-informatics ~\citep{gilmer2017neural,zhang2018end}, recommender systems ~\citep{ying2018graph, wang2019knowledge,fan2018learning}, question-answering systems~\citep{schlichtkrull2018modeling} and combinatorial problems~\citep{khalil2017learning,li2018combinatorial, gasse2019exact}. The popular examples include variants of graph neural networks (GNNs)~\citep{gori2005new,scarselli2008graph} and receptive field embeddings~\citep{grover2016node2vec,wang2016structural}.

The versatile applications of GRL algorithms motivate the theoretical study that we present in this work. In the structured learning problem, we are given a graph $\ca G$ containing $N$ vertices with vertex set $\ca V$ and edge set $\ca E$. All edges are assumed to be unordered. As in the typical supervised learning problem, we assume that vertex indexed by $i$ has its feature $\bfa X_i$ and label $\bfa Y_i$, which can be grouped as $\bfa Z_i$. The goal is to infer $\bfa Y_i$ of all vertices. GRL algorithms address this problem through the localized view. For each vertex $i$, we can construct its \textit{receptive field}, denoted by $\Xi(i)$. The returning hypothesis of GRL is is a hypothesis $h$ that maps the feature in the receptive field $\{\bfa X_j:j\in\Xi(i)\}$ to $\bfa Y_i$. As the receptive field is handcrafted, one can construct it based on the naturally present edges or even use a randomized strategy. For simplicity, we limit the discussion to the fixed receptive field in this work.

This supervised learning problem also requires us to introduce a statistical model of the random features in the network. Contrary to the standard setting of statistical learning, where the data is assumed to be i.i.d. sampled from some unknown distribution, the vertices in the graph cannot be considered as i.i.d. Otherwise, the structure of the graph is invalidated, and the problem becomes degenerated. To address this limitation, we introduce Dobrushin's condition, which has been proved to induce concentration of measure~\citep{kulske2003concentration} while allowing statistical dependencies across node features. 



In order to provide learning guarantees, in this work, we focus on the notion of uniform algorithmic stability, first introduced by Bousquet and Elisseeff \citep{bousquet2002stability}. 
Uniform stability ~\citep{feldman2018generalization,feldman2019high} was shown to be a favorable choice of sufficient condition for the consistency of hypothesis returned by learning algorithms. Concretely, we propose a framework of multi-fidelity stability, a relaxation from the previous ones on the non-i.i.d. data. As the topology of the graph comes into play, we introduce two different types of stability based on the relative position of vertices in the receptive field of the targeted vertex. The difference between the two types is termed \textit{ discrepancy } in this work. The multi-fidelity stability generalizes the standard setup to the learnability of GRL algorithms. We also demonstrate that the gap between the two types of stability is remarkable in the equivariant models by obtaining a lower bound on the discrepancy.

Within the framework of uniform stability, we can have a consistent hypothesis when the algorithms have $O(\min\{\frac{1}{\sqrt N},\frac{1}{\bar d}\})$ stability where $\bar d$ is the average degree and $N$ is the number of training nodes. This implies that the algorithm can be consistent with the size of the graph, or \textit{generalize in a single large graph}. The phenomenon of generalization with a single graph has also been studied in~\citet{london2016stability} under the standard point-to-point structured learning problem. We find three necessary conditions for this phenomenon: \textbf{(1)} The dependence of measure between adjacent vertices. \textbf{(2)}. The asymptotic stability of the algorithm. \textbf{(3)}. The sparsity of the receptive field constructed via GRL algorithms. The first and the third condition is novel.

To demonstrate the applicability of this framework in real problems, we case study two typical algorithms: \textbf{(1)}: SGD in the convex and non-convex landscape. 
\textbf{(2)}: Constrained regression on equivariant neural network. We give upper bounds for the uniform stability, which depends on the sparsity of the model as follows: For SGD in the convex landscape, the hypothesis is consistent when the maximum degree being $O(N^{\frac{1}{4}})$. While in a non-convex landscape, the model needs to be extremely sparse (i.e., the maximum degree is $O(1)$ ). 

The primary advancement obtained in this work are summarized as follows:
\begin{compactenum}
    \item To the best of our knowledge, this work is the first one to give the generalization guarantees of GRL for structured prediction. Our result suggests that a sparsely structured and stable GRL algorithm guarantees the generalization in the number of vertices.  
    \item We introduce multi-fidelity stability, a weakened condition than uniform stability, by addressing the in-the-set vertices and out-of-set vertices separately. We also provide a lower bound for their difference on a 1-layer equivariant GNN as a justification.
    \item We obtain a high probability upper bound on the multi-fidelity stability for algorithms trained with SGD. Our method is built upon the technique of ~\citet{hardt2016train}'s work to estimate the uniform stability, but we give non-asymptotic bound for the generalization gap while theirs only presented first moment bound. 
    
\end{compactenum}

\section{Related Work}
Several previous works have been attempted to address the generalization bound of GNNs on classifying i.i.d. generated graphs ~\citep{garg2020generalization,liao2020pac} In those works, graphs are assumed to be i.i.d. sampled from some random graph models. Their work is limited to the hypothesis class of graph neural networks. On the contrary, this work studies the learnability of structured prediction tasks with GRL, which requires us to go beyond the i.i.d. data assumption. Our bound can potentially be transferred to these algorithms,
which might be a promising future direction to explore.

Another related line of work is the standard literature of learnability in structured prediction.
A line of work summarized in~\citet{london2016stability} utilizes the concentration inequality obtained through martingale methods~\citep{kontorovich2008concentration} to study the common structured prediction problems, but their work studied the point-to-point regime, which is a special case of GRL, where only the target itself presents in the receptive field. They introduced different definitions of stability, the maximal difference of output, which differs from the algorithmic stability in this work. They also claimed that even with only a single large graph, an algorithm could generalize.  The result presented in this work also attests to such observation's validity in the hypothesis returned by GRL algorithms, but we argue that it is only possible when sparsity presents. Other earlier work focuses on several parametric models like conditional Markov networks ~\citep{roller2004max,ando2005framework,wigler2013revascularization,bradley2012sample} are less correlated to this work since they assume the parametric models explicitly while we only specify the model as having a weak dependency.

A few practical works consider the transduction of GRL from subgraphs.~\citep{hamilton2017inductive,zeng2019graphsaint}. A line of theoretical work also considers the generalization guarantees for transduction on i.i.d. samples~\citep{cortes2007transductive,el2009transductive,el2006stable}. In transduction, researchers aim to study the generalization from a subgraph to a whole one. Our discussion does not consider this problem and centers on inductive learnability. However, this might become a promising direction for future work.
\section{Preliminaries}


\subsection{Problem Formulation}
In this work, we assume that the graph $\ca G(\ca V,\ca E)$ with $N$ vertices has index set $\ca V=\{1,\ldots,N\}$, an unordered edge set $\ca E$ and the corresponding adjacency matrix $\bfa A\in\bb R^{N\times N}$. Vertex are indexed by $i$ in $\ca G$ and have a random feature vector $\bfa X_i\in\bb X\subset\bb R^d$ as well as a label $\bfa Y_i\in\bb Y\subset\bb R$. We group them together as $\bfa Z_i=(\bfa X_i,\bfa Y_i)\in\bb Z$. We assume that $\bfa Z_i$ is drawn according to some distribution $\bb D$. For simplicity, we denote $\bfa X_i^j=(\bfa X_i,...,\bfa X_j)$ , $\bfa Y_i^j=(\bfa Y_i,...,\bfa Y_j)$ and $\ca Z:=\bfa Z_1^N=(\bfa Z_1,\ldots,\bfa Z_N)$. 

We let $\ca P(\ca V)$ be the power set of $\ca V$ and define $\ca H \subset \{h:\bb X^{\ca P(\ca V)}\rightarrow \bb Y\} $ to be the hypothesis set. We consider two learning setups:
\begin{compactenum}
    \item  We have a single large graph where $\ca Z$ is the joint feature/label pair of vertices in the graph. The learning algorithm $\ca A:\bb Z^N\rightarrow \ca H$ returns a hypothesis $h_{\ca Z}\in\ca H$ after taking all feature/label pairs in a graph as input.
    \item We have $m$ graphs generated i.i.d. according to $\bb D$, where $m$ batches of vertices are assembled as $\ca Z_1^m$. The algorithm takes as input $m$ batches of vertices and returns a hypothesis $h_{\ca Z_1^m}$ upon optimizing over all vertices. We denote $\bfa Z_i^{(j)}$ as the feature label pair of the vertex $i$ of $\ca Z_j$ and let $S_i^{(j)}=(\ca T_i^{(j)},\bfa Y_i^{(j)})$ be the feature set of receptive field indices and label of vertex $i$ of $\ca Z_j$. 
\end{compactenum}

For the discretionary receptive field set $\Xi(i)$, we assume that $i\in\Xi(i)$ and $i\in\Xi(j)\Leftrightarrow j\in\Xi(i)$. One example for such receptive field set is the 1-hop neighborhood of vertices in the graph. This receptive field construction resembles the standard 1 layer GNN.  We augment the feature of a vertex by its receptive field by defining $\ca T_i=\{\bfa X_j: j \in \Xi(i)\} $. Hence, any hypothesis $h$ takes as input a set $\ca T_i$ and makes a prediction for the label of $i$-th vertex by $h(\ca T_i)$. We also group together this augmented feature set with the label as $S_{i}=(\ca T_i,\bfa Y_i)$, $S_i^j=(S_i,...,S_j)$, and $\ca S = S_1^N\in \bb S$ with $\bb S=\{\ca S:\ca S \text{ is induced by }\ca Z\text{ with }\ca Z\in\bb Z^{N}\}$. Recall that we denote $card( T_i)=\ca N_i$, we denote the normalized sparsity as $d_i=\frac{\ca N_i}{N}$ and let $\bar d=\sum_{i=1}^N d_i$ be the average sparsity of the receptive field.


\subsection{Notations in Learning Theory} 
We review some standard notations in learning theory. Let $L:\bb Y\times\bb Y\rightarrow\bb R^+$ denote the loss function. Without further specification in the context, we assume that this loss is uniformly bounded: $L(\wh y,y)\leq B_L$. For the two different settings stated previously: In \textbf{(1)}, we denote $\wh R_{\ca Z}(h)=\frac{1}{N} \sum_{i\in\ca V}L(h(\ca T_i),\bfa Y_i)$ as the empirical error on the training set of samples $\ca Z$.
In \textbf{(2)}, we denote $\wh R_{\ca Z_1^{m}}(h)=\sum_{j=1}^m \frac{1}{Nm} \sum_{i\in\ca V}L(h(\ca T_i^{(j)}),\bfa Y_i^{(j)})$ as the average empirical error over a training set containing $m$ sets of samples $\ca Z_1^m=(\ca Z_1,...,\ca Z_m)$ drawn i.i.d. from $\ca G$ according to $\bb D$. We let $R(h)$ be the generalization error of $h$ defined by $R(h)=\bb E_{\ca Z\sim\bb D}[\wh R_{\ca Z}(h)|h]$ that is the conditional expectation on some random/fixed hypothesis $h$.

According to standard notation in probability, we use capital letters like $\bfa Z$ to denote random variables or a set of variables and lower case letters like $\bfa z$ to denote their values.


\subsection{Weak Dependency}
We assume that $\bb D=\bfa P(\bfa Z_1^N)$ satisfies the Dubrushin's uniqueness condition (Dubrushin's condition for short) stated as follows:
\begin{definition}[Dobrushin's uniqueness condition]
Let $\bfa Z_1^N=(\bfa Z_1,...,\bfa Z_N)$ be a random vector over $\bb Z^N$.Let $\bfa Z_{-i-j}$ denote $\bfa Z_1^N\setminus\{\bfa Z_i,\bfa Z_j\}$ and $\bfa z_{-i-j}$ similarly. For $i\neq j$, $i,j \in\{1,...,N\}$ define 
\begin{align*}
    I_{i,j}(\bfa Z_1^N)&=\sup_{\substack{\bfa z_{-i-j}\in \bb Z^{N-2}\bfa z_j,\bfa z_j^\prime\in\bb Z}}\\
    TV(&{\bfa P(\bfa Z_i|\bfa Z_{-i-j}=\bfa z_{-i-j},\bfa Z_j=\bfa z_j)}, \\
    &{\bfa P(\bfa Z_i|\bfa Z_{-i-j}=\bfa z_{-i-j},\bfa Z_j=\bfa z_j^\prime)})
\end{align*}
We say that the vector $\bfa Z_1^N$ satisfies Dobrushin's uniqueness condition with coefficient $\alpha$ if $\sup_{i,j}I_{i,j}(\bfa Z_1^N)= \alpha  \leq 1$.
\end{definition}
Dobrushin's condition implies empirical measure concentration in the weakly dependent sets~\citep{kulske2003concentration} .
This condition implies that a single vertex in the graph will be only weakly dependent on any vertices in its receptive field.

\subsection{Discussion}
The two different setups result from the fundamental properties of structured prediction. Under the regularity of weak dependency, the algorithm can generalize to unseen graphs even with a single sample, as long as it is sufficiently large. This property of generalization uniquely presents in GRL. On the contrary, we are also interested in the standard setup, where the weak law of large numbers guarantees that the average of a sufficient number of i.i.d. samples converges in probability to their expectation. 

\section{Multi-fidelity Stability}

This section introduces multi-fidelity stability, followed by the generalization guarantees obtained for multi-fidelity and uniform stable GRL algorithms in the $1$-graph and $m$-graphs learning settings.

Algorithms with uniform stability ~\citep{bousquet2002stability} generalize as their deviation is bounded when trained with two sets of samples differing in a single sample. We further take into account the topology naturally induced by GRL in this framework. Multi-fidelity stability is a weaker condition than uniform stability is discussed in this section.

\begin{definition}[Multi-fidelity Stability]
Given $\ca G(\ca V,\ca E)$, let $\ca Z$ and $\ca Z^i$ be any two sets of samples drawn from $\bb  Z^N$ but differing by a single vertex $i$'s feature and label (i.e. $\ca Z=(\ca Z^i\setminus \{\bfa Z_i\})\cup \{\bfa Z_i^\prime\}$ ). We denote by $h_{\ca Z}$ and $h_{\ca Z^i}$ the hypothesis returned by learning algorithm $\ca A$ when trained on $\ca Z$ and $\ca Z^i$ respectively. Then the algorithm $\ca A$ is said to have $i$-th type-1 stability $\beta_{1,i}$ concerning the loss function $L$ if the hypotheses it returns when trained on any such samples $\ca Z$,$\ca Z^i$ satisfy:
\begin{equation*}
\sup_{\ca S^\prime\in\bb S }\sup_{j : j\notin\Xi(i)}[ |L(h_{\ca Z}(\ca T^\prime_j), \bfa Y^\prime_j) -L(h_{\ca Z^i}(\ca T^\prime_j),\bfa Y^\prime_j)|]= \beta_{1,i}
\end{equation*}
with $S^\prime_j=(\ca T^\prime_j,\bfa Y^\prime_j)\in \ca S^\prime$ being the $j$-th augmented feature set in $\ca S^\prime$.
Additionally, it has $i$-th type-2 uniform stability $\beta_{2,i}$ (or $i$-th uniform stability for short)  with respect to the loss function $L$ if the hypotheses it returns when trained on any such samples $\ca Z$,$\ca Z^i$ satisfy:
\begin{equation*}
\sup_{\ca S^\prime\in\bb S }\sup_{j\in\ca V }[ |L(h_{\ca Z}(\ca T^\prime_j), \bfa Y^\prime_j) -L(h_{\ca Z^i}(\ca T^\prime_j),\bfa Y^\prime_j)|]= \beta_{2,i}
\end{equation*}
with $S^\prime_j=(\ca T^\prime_j,\bfa Y^\prime_j)\in \ca S^\prime$.
Moreover, we say that it has type-1 stability $\beta_1$ and type-2 stability $\beta_2$ if:
\begin{equation*}
    \sup_{i\in\ca V}\beta_{1,i}=\beta_1,\;\;\;\;\sup_{i\in\ca V}\beta_{2,i}=\beta_2
\end{equation*}
One can see that $\beta_1\leq\beta_2$ by definition. Whence we denote $\beta_2-\beta_1$ as the \textbf{discrepancy} of algorithm $\ca A$.
\end{definition}
In the multi-fidelity stability, we distinguish between vertices within/without the receptive field. This characteristic differs from the standard algorithmic stability where vertices are treated equally.



We also define uniform stability in the multi-graph regime. It guarantees the generalization when algorithm $\ca A$ takes multiple sets of samples from $\bb D$ as the training set.
\begin{definition}[Uniform Stability]
\label{def2}
Given $\ca G(\ca V,\ca E)$, let $\ca Z_1^m$ and $\ca Z_1^{\prime m}$ be any two $m$-sized sets of samples drawn i.i.d. from $\ca G$ according to $\bb D$ but differ by an item $\bfa Z_i^{(j)}\in\ca Z_j$ (i.e.$\ca Z_{j}^\prime=\ca Z_j\setminus\{\bfa Z_i^{(j)}\}\cup\{\bfa Z_i^{\prime (j)}\}$ ). Then the algorithm $\ca A$ is said to have $\mu$ uniform stability (or is $\mu$-uniform stable) with respect to the loss function $L$ if the hypothesis it returns when trained on any such samples $\ca Z_1^m,\ca Z_1^{\prime m}$ and for any sample size $m$ satisfy:
\begin{equation*}
    \sup_{m\geq 1}\sup_{\ca S^\prime\in\bb S}\sup_{i,j\in\ca V} [ |L(h_{\ca Z_1^m}(\ca T_j^\prime),\bfa Y_j^\prime)-L(h_{\ca Z_1^{\prime m}}(\ca T_j^\prime),\bfa Y_j^\prime)|]=\mu
\end{equation*} with $S_j^\prime=(\ca T_j^\prime,\bfa Y_j^\prime)\in\ca S^\prime$.
By definition, we have $\beta_1\leq\beta_2\leq\mu$.
\end{definition}

With those two types of stability at hand, we can formally establish the guarantees of learning with a single large graph as follows:
\begin{theorem}[Single Graph Generalization]\label{thm:2}
\label{twoo}Given $\ca G(\ca V,\ca E)$, assume that the loss function $L$ is upper bounded by $B_L\geq 0$. Let $\ca A$ be a learning algorithm with type-1 stability $\beta_1$ and type-2 stability $\beta_2$ and let $\ca Z$ be a single set of samples drawn from $\ca G$ according to $\bb D$. Assume that $\bfa P(\bfa Z_1^N)$ satisfies Dobrushin's condition with coefficient $\alpha$. 
Let $d_i=\frac{\ca N_i}{N}$ and $\Xi(i)$ be the neighborhood index set of $i$ with $\ca N_i=card(\Xi(i))$ and $N=card(\ca V)$.  Then, for all $\delta\in(0,1)$, with probability at least $1-\delta$ over $\ca Z$ drawn, the following holds:
\begin{align*}
   &R(h_{\ca Z})\leq \wh R_{\ca Z}(h_{\ca Z})+2\bar d\beta_2\\
    &+\sqrt{2\sum_{i=1}^N((2-2d_i)\beta_1+d_i(\beta_2+B_{L}))^2}\sqrt{\frac{\log(1/\delta)}{1-\alpha}}
\end{align*}
\end{theorem}
\begin{remark}
We note that the generalization gap can be upper bound with two terms. The first one is the product of average degree with the type-2 stability, which gives a worst-case estimate on the expected generalization gap. The second term comes from the tail, where we see that when $\beta_1=\beta_2$ and let $d_i=\frac{1}{N}$, this bound will degrade to the classical one obtained by ~\citet{bousquet2002stability}. It is also important to note that this term primarily depends on $\beta_1$. Then generalization in vertices holds as long as our algorithm is with $\beta_1\leq \beta_2= O(\min(1/{\sqrt{N}},1/{\bar d}))$. 
\end{remark}
  
The generalization to multiple graphs leads to the following guarantee:
\begin{theorem}[$m$-Graphs Generalization]
\label{thm:3}
With the  notations of Theorem \ref{twoo}, assume that we draw $\ca Z_1^m= \{\ca Z_1,...,\ca Z_m\}$ from $\bb Z^N$ i.i.d. according to $\bb D$, and we let $h_{\ca Z_1^m}$ be the hypothesis returned by $\mu$-uniform stable algorithm $\ca A$ when trained on $\ca Z_1^m$. Then, for all $\delta\in(0,1)$, with probability at least $1-\delta$ over $\ca Z_1^m$ drawn, the following holds:
\begin{align*}
    &\emrr{\ca Z_1^m} \leq\mrr{\ca Z_1^m}+ N\mu\\
    &+\sqrt{2m\sum_{i=1}^N((2-\frac{d_i}{m})\mu+\frac{d_iB_L}{m})^2}\sqrt{\frac{\log(1/\delta)}{1-\alpha}}.
\end{align*}
\end{theorem}
\begin{remark}
Generalization guarantees for learning $m$ graphs with stable GRL algorithm is similar to the i.i.d. problem, where $O(1/{\sqrt{m}})$ stability suffices for consistency. However, the major limitation of this bound lies in the second term on the R.H.S., which depends linearly on $N$. A line of recent work devoted to sharpening the generalization bound of uniform stable algorithm ~\citep{feldman2018generalization,feldman2019high} can be used to sharpen this bound, but we omit it here.
\end{remark}

\section{Stability Estimation}
In this section, we give upper-bounds on multi-fidelity stability of two standard algorithms:\begin{compactenum}
    \item Stochastic gradient descent (SGD) in smooth convex/non-convex landscapes, where our result shows that sparse receptive field generalizes well and that the stability in the non-convex case has an asymptotically worse rate than the convex case. The method follows ~\citet{hardt2016train}, but we provide a non-asymptotic upper bound, which improves their result.
    \item The 1-layer equivariant 
GNN. In this example, we justify that the \textit{discrepancy} plays a key part in the equivariant models via a lower bound. 
\end{compactenum}

\subsection{Stochastic Gradient Descent}
Recent models like GNNs are normally optimized with the first-order stochastic optimization methods. 

In the standard setting, the update rule for SGD can be roughly formalized as:
\begin{align*}
    \bfa w_{t+1}=\bfa w_t-\alpha_t\nabla_{\bfa w_t} f(\bfa w_t,\bfa x_t)
\end{align*} where $\bfa w_t$ is the weight vector at round $t$, $\nabla_{\bfa w_t} f$ is the gradient of objective function $f$ w.r.t. $\bfa w_t$, $\alpha_t$ is the step-size at round $t$ and $\bfa x_t$ is the data related argument of function. $f$ is treated as a black box function without closed form.

SGD can naturally be extended to become a GRL algorithm. We denote $\fk S_i^j=(\{\bfa Z_k:k\in\Xi(i)\}\setminus \{\bfa Z_j\})\cup\{\bfa Z_j^\prime\}$ to be set of nodes in the neighborhood of $i$-th node with some $\bfa Z_j$ (assuming that $j\in\Xi(i)$) replaced by $\bfa Z_j^\prime$.

We denote the objective as $f(\fk S_i,\bfa w)=L(h_{\bfa w}(\ca T_i),\bfa Y_i)$ with $\fk S_i =(\ca T_i,\bfa Y_i)$ and $h_{\bfa w}$ being the hypothesis parameterized by $\bfa w$. Assuming that in this case, each time an index $i$ is randomly picked from $[N]$ and the update rule will became
$$
\bfa w_{t+1}=\bfa w_t-\alpha\nabla_{\bfa w} f(\fk S_i,\bfa w_t)
$$

Introducing the function $$G(\bfa w,\alpha, i)=\bfa w-\alpha\nabla_{\bfa w} f(\fk S_i,\bfa w)$$ we can rewrite the update function as $$\bfa w_{t+1}=G(\bfa w_{t},\alpha_t,i)$$ for simplicity. We use the operator $\delta^i $ that acts on any real value/vector/function $K$ (e.g. $\delta^iK = K-K^i$ ) to denote the difference between the value/vector/function $K$ returned by SGD algorithm when trained with two sets of samples $\ca Z$, $\ca Z^i$ drawn from $\ca G$ that differ in $\bfa Z_i$. For example, $\delta^i\bfa w_t=\bfa w_t-\bfa w_t^i$ is the difference between the weight vector returned at $t$-th round when we replace $\ca Z$ with $\ca Z^i$ as training set. In what follows, we also denote $\Vert\cdot\Vert_2$ as $\Vert\cdot\Vert$.

Note that $\beta_1\leq\beta_2$. Hence, the hypothesis's generalization will be guaranteed when we obtain a proper upper bound for $\beta_2$.
\subsubsection{Assumptions:} To formulate our discussion, we make the following common assumptions in the convex/non-convex setting:
\begin{compactenum}
    \item Smoothness: $f$ is $\lambda$-smooth with respect to $\bfa w$.
    \item We assumed that the diameter of $\bb Z$ is upper-bounded, (e.g. $\sup_{i,j}\Vert\bfa Z_i-\bfa Z_j\Vert=B_Z$)
    \item Lipschitzness: For all $i\in\ca V$ and $\bfa w,\bfa w^\prime\in\bb W$: $f(\tz i,\bfa w)-f(\tz i,\bfa w^\prime)\leq\ca L\Vert\bfa w-\bfa w^\prime \Vert$ 
    \item Gradient Lipschitzness: $\Vert \nabla_{\bfa w}f(\tz{i},\bfa w)-\nabla_{\bfa w}f(\dtz{j}{i},\bfa w)\Vert\leq\zeta\Vert \bfa Z_j-\bfa Z^\prime_j\Vert$
\end{compactenum}

\subsection{$\gamma$-Strongly convex regime}
In the strongly convex regime, we can upper bound the first moment of type 2 stability of vertex indexed by $i$: 
\begin{lemma}[First moment bound with convexity ]
\label{lm:6}
\label{one}
Assume that $f(\tz{i},\bfa w)$ is $\lambda$-smooth and $\gamma$-strongly convex. Suppose we run SGD with fixed step size $\alpha$ s.t. $\alpha^4\lambda^2+\frac{2\alpha\lambda\gamma}{\lambda+\gamma}\leq 1$. The algorithm induced by $T$-step SGD has expected $i$-th type-2 stability and type-2 stability upperbounded by:
\begin{align*}
    &\bb E[\beta_{2,i}]\leq \ca L(\pa Z_{i}^T-1)\frac{\pa Y_{i}}{\pa Z_{i}-1}\;,\;\bb E[\beta_2]= O(\sup_i d_i)~,
 \end{align*}
with 
$   \pa Z_{i}=d_i\alpha\lambda(\frac{\gamma}{\lambda+\gamma}-\alpha)+\frac{\alpha^2\lambda}{N}+(1-\frac{\alpha\lambda\gamma}{\lambda+\gamma})$$  and $$
    \pa Y_{i}=\alpha B\zeta\frac{\ca N_i-1}{N}+2\alpha\ca L\frac{1}{N}
$\end{lemma}

\begin{remark}
When $\pa Z_{i}\leq 1$ for all $i$, the expected type-2 stability will converge as $T\rightarrow\infty$. To meet this, we either need to choose a small step size, or our function is very smooth. It is not observed in the i.i.d. case and is particularly induced by the GRL algorithm.
\end{remark}
However, the above result only indicates generalization guarantees in expectation. Our result further extends to  high probability bounds:
\begin{theorem}[Non-asymptotic bound with convexity]
\label{thm:8}
\label{two}
Under the same conditions of theorem \ref{one}, with probability at least $1-\delta$, the following holds:
\begin{align*}
&\beta_2\leq (\ca L+(\lambda-\gamma)\sqrt{\frac{\log\frac{2}{\delta}}{8}})\sup_i\bigg((\pa Z_i^T-1)\frac{\pa Y_i}{\pa Z_i-1}\bigg)\\
    &+(\lambda-\gamma)\sqrt{\frac{\log(\frac{2}{\delta})}{8}}\bigg(\sup_i\bigg[(\pa Z_i^T-1)\frac{\pa Y_i}{\pa Z_i-1}\bigg]\\&+\sqrt{\frac{1}{\delta}\sum_{i=1}^N\bigg( (2\pa Y_{i}^2)\frac{\pa Z_{i}^{2T}-\pa Z_{i}^T}{\pa Z_{i}^2-\pa Z_{i}}+\pa Y_{i}^2\frac{1-\pa Z_{i}^T}{(1-\pa Z_{i})^2}\bigg)}\bigg)^2\\
    &=O(\frac{\sup_i\ca N^2_i}{N})=O(\sup_id_i\ca N_i)
\end{align*}
\end{theorem}
\begin{remark}
To guarantee generalization almost surely at any number of steps, we need the maximum receptive field to have sparsity $\sup_i\ca N_i=O(N^{1/4})$. 
\end{remark}
The above theorem immediately yields a high probability upper bound on the generalization gap of convex and smooth SGD, which is stated as follows:
\begin{corollary}
\label{thm:9}
Assuming that $f(\tz{i},\bfa w)$ is $\lambda$-smooth and $\gamma$-strongly convex for all $i\in\ca V$, then with probability at least $1-\delta$ we have 
\begin{align*}
     R(h)&\leq\wh R(h)+\bigg[(2-\frac{1}{N})\sqrt{2N\log \frac{2}{\delta}}+2\bigg]\\
     &\cdot\sup_{i\in\ca V}\bigg[\ca L(\pa Z_{i}^T-1)\frac{\pa Y_{i}}{\pa Z_{i}-1}+\sqrt{\frac{1}{4\delta}}(\lambda-\gamma)\\
     &\bigg(\frac{4}{\delta}\bigg( (2\pa Y_{i}^2)\frac{\pa Z_{i}^{2T}-\pa Z_{i}^T}{\pa Z_{i}^2-\pa Z_{i}}+\pa Y_{i}^2\frac{1-\pa Z_{i}^T}{(1-\pa Z_{i})^2}\bigg)\bigg)\bigg]\\
     &+\frac{B_L}{N}\sqrt{2N\log\frac{2}{\delta}}~.
\end{align*}
\end{corollary}
However, strongly convex assumptions is overly strict and can be un realistic in the real cases. Most algorithms in GRL suffered from the non-convexity. This poses great challenges to the learnability as well ~\citep{hardt2016train}. We then move on to the general case where no convexity presents. In this problem, we can observe a significant degrade of generalization guarantees. 
\subsection{Non-convex regime}
In the non-convex regime, our result suggests that the generalization guarantees will be compromised. In particular, the asymptotic rate argues for stringent condition on the receptive field size of $O(1)$. Intuitively, this suggests that GRL will have to only capture very sparse local structure instead.
\begin{lemma}[First moment without convexity]
\label{lm:10}Assume that $f(\tz i,\bfa w)$ is $\lambda$-smooth. Then the algorithm induced by running SGD $T$-steps with fixed step size $\alpha$ has the expected $i$-th type-2 stability and type-2 stability upperbounded by:
\begin{align*}
\bb E[\beta_{2,i}]&\leq\ca L (\pa M^T-1)\frac{\pa Y_{i}}{\pa M-1}, \\
    \bb E[\beta_2]&=\ca L(\pa M^T-1)\frac{\sup_i\pa Y_i}{\pa M-1}=O(\sup_i d_i)
\end{align*}
with
$
    \pa M=\frac{N-1}{N}\alpha\lambda
$ and $\pa Y_i=\alpha B_Z\zeta\frac{\ca N_i-1}{N}+2\alpha\ca L\frac{1}{N}$.
\end{lemma}
The first moment of stability follows the method developed in ~\citet{hardt2016train}. We further turn this result into a high probability upper bound, similarly as in the convex case. This has also been discussed in ~\citet{feldman2019high}:



\begin{theorem}[Non-asymptotic bound without convexity]
Assume that $f(\tz{i},\bfa w)$ is $\lambda$-smooth and non-convex with the same conditions and notations in lemma \ref{lm:10}. Then the following holds with probability at least $1-\delta$ over $\ca Z$ drawn:
    \begin{align*}
    &\beta_2\leq \ca L (\pa M_t^T-1)\frac{\sup_i\pa Y_i}{\pa M_t-1}\bigg(1+\sqrt{\frac{\log{\frac{2}{\delta}}}{2}}\bigg)+\\
    &\ca L\sqrt{\frac{\log\frac{2}{\delta}}{\delta}\sum_{i=1}^N\bigg((2\pa Y_{i}^2)\frac{\pa M_{t}^{2T}-\pa M_{t}^T}{\pa M_{t}^2-\pa M_{t}}+\pa Y_{i}^2\frac{1-\pa M_{t}^T}{(1-\pa M_{t})^2}\bigg)}\\
    &= O\bigg(\frac{(\sup_i\ca N_i)^2}{\sqrt{N}}\bigg)~.
\end{align*}
\end{theorem}
\begin{remark}
To ensure the $O(\frac{1}{\sqrt{N}})$ type-2 stability, we need $\sup_i\ca N_i=O(1)$ as $N\rightarrow\infty$. This condition will make the receptive field significantly sparser than the convex case, where we only need $\sup_i\ca N_i=O(N^{1/4})$. To ensure convergence in $T$, we will need $\pa M=\frac{N-1}{N}\alpha\lambda\leq 1$.
\end{remark}
The following corollary gives the high probability generalization bound for the non-convex regime:
\begin{corollary}[Generalization of non-convex optimization]
Assume that $f(\fk S_i,\bfa w)$ is $\lambda$-smooth and non-convex. With the assumptions stated before, and step size $\alpha_t=\alpha$ for all $t$, the following holds with probability at least $1-\delta$:
\begin{align*}
    R(h)&\leq\wh R(h)+\bigg[(2-\frac{1}{N})\sqrt{2N\log \frac{2}{\delta}}+2\bigg]\\
    &\cdot\sup_{i\in\ca V}\bigg [ \ca L (\pa M_{t}^T-1)\frac{\pa Y_{i}}{\pa M_{t}-1}\bigg(1+\sqrt{\frac{1}{\delta}}\bigg)\\
    &+\sqrt{\frac{4}{\delta}\bigg((2\pa Y_{i}^2)\frac{\pa M_{t}^{2T}-\pa M_{t}^T}{\pa M_{t}^2-\pa M_{t}}+\pa Y_{i}^2\frac{1-\pa M_{t}^T}{(1-\pa M_{t})^2}\bigg)}\bigg]\\
    &+\frac{B_L}{N}\sqrt{2N\log\frac{2}{\delta}}
\end{align*}
\end{corollary}

\textbf{Discussion}: The bound listed in this work relies on the necessary restraint that the uniform stability needs to scale $O(\frac{1}{\sqrt{N}})$. A recent line of work by ~\citet{feldman2018generalization,feldman2019high} gave a sharper tail for the uniform stable algorithms. Their work suggests that $O\big (\frac{1}{\log n}\big )$ uniform stability is enough for generalization. However, a non-trivial discussion will be needed to apply their technique to this work, and we leave it for future work.
   

\subsection{Graph Neural Networks}
Most graph neural networks used today treat the vertices in the receptive field heterogeneously. For example, when we know that $A,B$ are connected to $C$, they can contribute to $C$ in different magnitude. Therefore, the algorithm addresses them by heterogeneous weight. We studied a simple equivariant model here, which consists of a single layer GNN. The main result obtained in this section is a sharp upper bound for the type-2 stability and a lower bound for discrepancy under two typical perturbation themes \begin{compactenum}
    \item label perturbation  
    \item first-order feature perturbation
\end{compactenum} They justified that our multi-fidelity design is non-trivial.

Here we simplify the feature set $\bfa X_1^N$ is concatenated to be a matrix $\bfa X\in\bb X^N\subset\bb R^{N\times m}$ and label set $\bfa Y_1^N$ concatenated as $\bfa y\in\bb Y^{N\times 1}\subset\bb R^{N\times 1}$ . Denote $\bfa Z=(\bfa X,\bfa y)$. Assume that $\bfa w\in\bb R^{m}$ is a fixed parameter. And the estimate is written as 
\begin{align*}
    \wh{\bfa y}=\tda\bfa X\bfa w
\end{align*}
with $\tda\in\bb A$ being the weighted adjacency matrix with $\bb A=\{\bfa B:\bfa B=\bfa B^\top\}\subset\bb R^{N\times N}$. Assume that $\Vert\bfa X_j\Vert_2\leq B_X$ for all $j\in\ca V$. $\Vert\bfa y\Vert_\infty\leq B_y$, and $\Vert\bfa w\Vert_2\leq B_w$. Denote $\otimes$ by the element-wise product. We can rewrite the network in the entry-wise form
\begin{align*}
   \wh y_i= h(\ca T_i)=\sum_{j\in\Xi(i)}\tda_{i,j}\bfa X_j\bfa w~.
\end{align*}

Using the regularized MSE as the loss function leads to the following objective:
\begin{align*}
    \underset{\tda\in\bb A,\text{ s.t. }\Vert\tda\Vert_F\leq C}{\text{ minimize }}\Vert\bfa y-\tda\bfa X\bfa w\Vert_2^2
\end{align*}
Instead of dealing with the constrained optimization problem, we introduced a slack variable $\gamma$ as the regularization parameter.
\begin{align*}
\label{mini}
   \underset{\tda\in\bb A}{\text{ minimize }}f(\tda )&= \underset{\tda\in\bb A}{\text{ minimize }}\Vert\bfa y-\tda\bfa X\bfa w\Vert_2^2\\&+\frac{\gamma}{2}\Vert\tda\Vert_F^2\tag{+}
\end{align*}
with $\gamma\in\bb R^+$ being the parameter that controls the magnitude of regularization. This parameter is crucial, as is suggested by ~\citet{shalev2010learnability} that the stability of the algorithm depends on the parameter of regularization.
For any matrix $\tda$ returned by algorithms solving (\ref{mini}) on $\bfa Z^i$, we denote $\tda^i$ as the matrix returned by (\ref{mini})  on $\bfa Z^i=(\bfa X^i,\bfa y^i)$ such that $\bfa X$ and $\bfa X^i$, $\bfa y$ and $\bfa y^i$ only differ in their $i$-th row. We further denote $\Delta_i \bfa A=\bfa A^i-\bfa A$. We summarize the result in this subsection as:
\begin{theorem}[Sharp Bound for Type-2 Stability]
\label{thm:12}
The algorithm defined by \ref{mini} have type-2 stability $\beta_2=\Theta(\sup_{i}d_i)$ 
\end{theorem}

\begin{theorem}[Discrepancy Lower Bound]
\label{thm:13}
The discrepancy (e.g. $\beta_2-\beta_1$) of algorithm solving ~(\ref{mini}) has the following lowerbound:
\begin{compactenum}
    \item $\Omega(\inf_{i}d_i)$ in the first order feature perturbation. (e.g.$\bfa y^i=\bfa y$, $\bfa X^i\neq\bfa X$, and $\Vert\bfa X^i-\bfa X\Vert\ll\Vert\bfa X\Vert$)
    \item $\Omega(\frac{1}{N})$ in the label perturbation. (e.g. $\bfa y^i\neq\bfa y$ and $\bfa X^i=\bfa X$)
\end{compactenum}
\end{theorem}
\begin{remark} The lower bound justified that multi-fidelity is non-trivial in GRL when the hypotheses are non-invariant w.r.t. the neighborhood set of the targeting vertex. A conjecture is that the discrepancy will only exist in algorithms returning non-invariant hypotheses. This discrepancy lower bound is also sharp (up to a constant factor) when the graph is sparse.
\end{remark}
 
In the SGD with the convex and non-convex setting, it is a non-trivial problem to obtain the lower bound for the discrepancy. In particular, we found that the optimization landscape crucially determines the asymptotic rate in the bound. Another crucial problem is the dependency on $T$. This can be treated with a proper choice of parameters in the convex case since the final optimal point is approachable. In the non-convex case, we will have exponential growth of stability parameters, which causes the bound to be vacuous when the number of timestamps becomes too large. This calls for a better estimate of stability for the non-convex regime.

The equivariant linear GNNs have a closed-form representation, but the closed-form stability is still unachievable yet. Our rate is based on the two extreme cases. If we are to perturb $\bfa X$ and $\bfa y$ simultaneously, the discrepancy will be hard to estimate. In practice, this 1-layer GNN is far too simple to address the problem posed by large datasets. This limitation restricts the applicability of the bound presented here. The new challenges induced by the topology of the receptive field in GRL algorithms left many explorable questions.



\section{Discussion}
\subsection{Limitations}
There are a few limitations in this work 
\begin{compactenum}
\item Most of the bound in this work relies on the quantity of the maximum receptive field, which could make the bound vacuous if the degree had a long tail. 
\item For the SGD in the non-convex case, our bound will scale almost exponentially with the number of timestamps, which is undesirable and vacuous when the number of timestamps is large. We believe a better method is needed to estimate the uniform stability under the non-convex landscape.

\end{compactenum}

\subsection{Open Problems and Future Work}
We give a summary of open problems together with several future directions of research in theory and applications.

\begin{compactenum}
\item We proved the lower bound of discrepancy in a special case. The case where the parameter $\bfa w$ is non-fixed is explorable.
\item Our study here is based on the algorithmic stability, some recent attempts have improved this framework through the methods used in adaptive and differentially private data analysis ~\citep{feldman2018generalization,feldman2019high}. We think it is possible to improve most of the results further by the idea in their technique. 
\item In the multi-graph learning problem, how to eliminate the $\Omega(N)$ term from the upper bound to achieve a generalization w.r.t. the scale of the graph is listed as an open problem.
\item We did not address the stability of algorithms that learn with multiple graphs as the training set, which might be of future interest.
\item The general perturbation lower bound in the equivariant example studied in this work.
\end{compactenum}



\section{Conclusion}
We analyzed the learning guarantees of graph representation learning by introducing a new measure of regularity on the GRL algorithm termed multi-fidelity stability. Our upper bound indicates that generalization in vertices, a novel phenomenon in GRL, depends highly on the sparsity of the receptive field of the algorithm. Moreover, our case study on SGD and equivariant single-layer GNN corroborate such a claim. Our lower bound on discrepancy justified that multi-fidelity stability is fundamental.

\section{Acknowledgement}
JB is partially supported by the Alfred P. Sloan Foundation, NSF RI-1816753, NSF CAREER CIF-1845360, and NSF CCF-1814524 and Samsung Electronics.
\bibliographystyle{plainnat}
\bibliography{bib.bib}
\section{Appendix}

\subsection{Literature of Convex Optimization}
The following common definitions of optimization literature are reviewed for completeness.
\begin{definition}[Convexity]
A function $f:\Omega\rightarrow\bb R$ is convex if for all $u,v\in\Omega$, we have$$
    f(u)- f(v)\geq\langle\nabla f(v),u-v\rangle
$$\end{definition}
\begin{definition}[Strongly Convex]
A function $f:\Omega\rightarrow\bb R$ is $\gamma$-strongly convex if for all $u,v\in\Omega$, we have $$
    f(u)-f(v)\geq\langle\nabla f(v),u-v\rangle +\frac{\gamma}{2}\Vert u-v\Vert^2
    $$
\end{definition}
\begin{definition}[Smoothness]
A function $f:\Omega\rightarrow\bb R$ is $\lambda$-smooth if for all $u,v\in\Omega$, we have$$
    \Vert \nabla f(u)-\nabla f(v)\Vert\leq \lambda\Vert u-v\Vert
 $$
    which is also equivalent to
$$
    f(u)-f(v)\leq\langle\nabla f(v),u-v\rangle+\frac{\lambda}{2}\Vert u-v\Vert^2
$$
\end{definition}
\begin{lemma}[Co-coerciveness]
\label{lm:co}
When the function $f$ is convex and $\lambda$-smooth, we have:
$$
    \langle\nabla f(v)-\nabla f(w),v-w\rangle\geq\frac{1}{\lambda}\Vert\nabla f(v)-\nabla f(w)\Vert^2
$$
\end{lemma}
The following lemmas are also used in the proof, which mainly due to ~\citet{nesterov2003introductory}
\begin{lemma}
\label{lm4}
Assume that $f$ is $\lambda$-smooth. Then the following properties hold:
\begin{compactenum}
    \item $G(\bfa w,\alpha,i)$ is $1+\alpha\lambda$ Lipschitz w.r.t $\bfa w$.
    \item If in addition $f$ is convex. Then for any $\alpha\leq\frac{2}{\lambda}$, $G(\bfa w,\alpha,i)$ is $1$-Lipschitz w.r.t. $\bfa w$.
    \item If $f$ is $\gamma$-strongly convex. Then for $\alpha\leq\frac{2}{\lambda+\gamma}$, $G(\bfa w,\alpha,i)$ is $(1-\frac{\alpha\lambda\gamma}{\lambda+\gamma})$ -Lipschitz w.r.t. $\bfa w$.
\end{compactenum}
\end{lemma}
\subsection{Proof of Theorem 1}
Dubrushin's condition leads to the concentration measure, which is obtained by~\citep{kulske2003concentration}. His result indicates that the upper bounds under the weakly dependent condition degrades only by a factor of $O(\frac{1}{\sqrt{1-\alpha}})$ than the upper bound in i.i.d. case. 
\begin{theorem*}
\label{th1}
    Assuming $\bb D$ is a distribution over $\bb Z^N$ satisfying the Dobrushin's condition with coefficient $\alpha$. Let $\bfa Z_1^N=(\bfa Z_1,...,\bfa Z_N)$ drawn according to $\bb D$ and $\Phi:\bb Z^N\rightarrow\bb R$ be a real valued function with the following property. 
    \begin{equation*}
        \forall \ca Z=\bfa Z_1^N,\ca Z^\prime=\bfa Z_1^{\prime N}\in\bb Z^N:\;\;\;|\Phi(\ca Z)-\Phi(\ca Z^\prime)|\leq\sum_{i=1}^N\bfa 1_{\bfa Z_i\neq\bfa Z_i^\prime}c_i
    \end{equation*}
    Then for all $t\geq 0$,
$
        \bfa P\bigg[f(\ca Z)-\bb E_{\bb D}[f(\ca Z)]\geq t \bigg]\leq \exp\bigg(-\frac{(1-\alpha)t^2}{2\sum_{i=1}^Nc_i^2}\bigg)
$
\end{theorem*}
Then we need the following lemma, which is a direct result of the definition of type-1 stability.
\begin{lemma}
\label{cor1.1}
Assuming algorithm $\ca A$ has type-2 stability $\beta_2$. Let $\ca Z$ be the set of sample drawn from $\ca G$ according to $\bb D$. We denote sample set that differ from $\ca Z$ in vertices indexed by elements in $\Lambda$ with $\Lambda=\{\Lambda_1,...,\Lambda_{card(\Lambda)}\}\subseteq \ca V$ as $\ca Z^{\Lambda}$.(e.g. $\ca Z^{\Lambda}=\ca Z\setminus\{\bfa Z_j:j\in\Lambda\}\cup\{\bfa Z^\prime_j:j\in\Lambda\}$). In particular, let $\ca Z^{\{ i\}}=\ca Z^i$. Let $\Lambda^j_1=\{\Lambda_1,...,\Lambda_j\}$ and in particular let $\Lambda=\Lambda_1^{card(\Lambda)}$ then for all $\ca Z$ and $\ca Z^{\Lambda}$ induced by $\ca Z$ over $\bb Z^N$, we have
\begin{equation*}
    \sup_{\ca S^\prime\in\bb S}\sup_{j\in\ca V}|L(h_{\ca Z}(\ca T^\prime_j),\bfa Y^\prime_j)-L(h_{\ca Z^{\Lambda}}(\ca T^\prime_j),\bfa Y^\prime_j)|\leq card(\Lambda)\cdot\beta_2 \;\;\text{ with } S^\prime_j=(\ca T^\prime_j,\bfa Y^\prime_j)\in \ca S^\prime
\end{equation*}
\end{lemma}
\vfill

Then we move to the proof of the theorem.

The proof is based on the concentration of measure inequality of the function $\Phi$ defined for all samples $\ca Z$ by $\Phi(\ca Z)=R(h_{\ca Z})-\wh R_{\ca Z}(h_{\ca Z})$. Let $\ca Z^i$ be another sample drawn from $\ca G$ according to $\bb D$ that differs from $\ca Z$ by $i$-th vertex, formally:
\begin{equation*}
    \ca Z = (\bfa Z_1,...,\bfa Z_{i-1},\bfa Z_i,\bfa Z_{i+1},...,\bfa Z_N),\;\;\ca Z^i = (\bfa Z_1,...,\bfa Z_{i-1},\bfa Z^\prime_i,\bfa Z_{i+1},...,\bfa Z_N)
\end{equation*}
Similarly, the $\ca S$, $\ca S^i$ induced by $\ca Z$,$\ca Z^i$ satisfy: 
\begin{equation*}
    \ca S=S_1^N\;\;\;\;\ca S^i=\ca S_1^i\cup S_2^i \text{ with } \ca S_1^i=\{S_j:i\notin \Xi(j)\}\text{ and }\ca S_2^i=\{S^\prime_j:i\in\Xi(j)\}
\end{equation*}
Then by definition of $\Phi$, the following inequality holds:
\begin{equation*}
    |\Phi(\ca Z^i)-\Phi(\ca Z)|\leq | R(h_{\ca Z^i})- R(h_{\ca Z})|+|\wh R_{\ca Z}(h_{\ca Z})-\wh R_{\ca Z_i}(h_{\ca Z^i})|
\end{equation*}
We then bound each of these two terms separately. By the type-1 $\beta_1$ stability and type-2 $\beta_2$ stability of $\ca A$ and the $B_L$ boundedness of $L$, we have:
\begin{align*}
    |\wh R_{\ca Z}(h_{\ca Z})-\wh R_{\ca Z^i}(h_{\ca Z^i})|&\leq\frac{1}{N}\sum_{S_j\in\ca S_1^i}|L(h_{\ca Z}(\ca T_j),\bfa Y_j)-L(h_{\ca Z^i}(\ca T_j),\bfa Y_j)| \\
    &+\frac{1}{N}\sum_{S_j\in\ca S_2^i}|L(h_{\ca Z}(\ca T_j),\bfa Y_j)-L(h_{\ca Z^i}(\ca T^\prime_j),\bfa Y^
    \prime_j)|\\
    &=\frac{(N-\ca N_i)\beta_1}{N}+\frac{\ca N_iB_L}{N}=(1-d_i)\beta_1+d_iB_L
\end{align*}

Also we immediately have that:
\begin{align*}
    |\mr{}-\mr{i}|&\leq\bb E_{\ca Z\sim\bb D}[\sum_{S_j\in\ca S_1^i}|L(h_{\ca Z}(\ca T_j),\bfa Y_j)-L(h_{\ca Z^i}(\ca T_j),\bfa Y_j)|\\
    &+\sum_{S_j\in\ca S_2^i}|L(h_{\ca Z}(\ca T_j),\bfa Y_j)-L(h_{\ca Z^i}(\ca T_j),\bfa Y_j)|]\\
    &\leq\frac{(N-\ca N_i)\beta_1+\ca N_i\beta_2}{N}=\beta_1+d_i(\beta_2-\beta_1)
\end{align*}
Hence
\begin{equation*}
   | \Phi(\ca Z^i)-\Phi(\ca Z)|\leq (2-2d_i)\beta_1+d_i(\beta_2+B_{L})
\end{equation*}
which gives that for all $\ca Z=\bfa Z_1^T,\ca Z^\prime=\bfa Z_1^{\prime T}\in\bb Z^N$
\begin{equation*}
    \big|\Phi(\ca Z)-\Phi(\ca Z^\prime)\big|\leq\sum_{i=1}^N\bfa 1_{\bfa Z_i\neq\bfa Z^\prime_i}\big((2-2d_i)\beta_1+d_i(\beta_2+B_{L})\big)
\end{equation*}
By theorem ~\ref{th1} we concluded that:
\begin{equation}
\label{arg}
    \bfa P(\Phi(\ca Z)-\bb E_{\bb D}\Phi(\ca Z)\geq \epsilon)\leq\exp\bigg(\frac{-(1-\alpha)\epsilon^2}{2\sum_{i=1}^N\big((2-2d_i)\beta_1+d_i(\beta_2+B_{L})\big)^2}\bigg)
\end{equation}
Then we move on to upper bound $\bb E_{\ca Z\sim\bb D}[\Phi(\ca Z)]$. 

We first note that by linearity of expectation 
\begin{align*}
    \bb E_{\ca Z\sim\bb D}\wh R_{\ca Z}(h_{\ca Z})=\frac{1}{N}\sum_{i\in\ca V}\bb E_{\ca Z\sim\bb D}L(h_{\ca Z}(\ca T_i),\bfa Y_i)\leq\frac{1}{N}\sum_{i\in\ca V}\bb E_{\ca Z\sim\bb D}[L(h_{\ca Z^{\Xi(i)}}(\ca T_i),\bfa Y_i)+\ca N_i\beta_2]\\
\end{align*}
where the inequality is given by replacing $\Lambda$ with $\Xi(i)$ in lemma ~\ref{cor1.1} for all $i\in\ca V$.
Hence
\begin{align*}
\label{to}
    \bb E_{\ca Z\sim\bb D}[\Phi(\ca Z)]&=\bb E_{\ca Z\sim\bb D}|R(h_{\ca Z})-\wh R_{\ca Z}(h_{\ca Z})|\\
    &\leq\frac{1}{N}\sum_{i\in\ca V}\bb E_{\ca Z\sim\bb D}[|L(h_{\ca Z}(\ca T_i),\bfa Y_i)-L(h_{\ca Z^{\Xi(i)}}(\ca T_i),\bfa Y_i)|+\ca N_i\beta_2]\\
    &\leq\frac{1}{N}\sum_{i\in\ca V}2\ca N_i\beta_2=2\bar d\beta_2
\end{align*}
In particular, this bound reduced to the i.i.d case in ~\citet{bousquet2002stability} when $\bar d=1$.
By replacing $\bb E_{\bb D}[\Phi(\ca Z)] $ in \ref{arg} and let its R.H.S. be $\delta$, we complete the proof.

\subsection{Proof of Theorem 2}
Similar to the one large graph case, our result depends on the following lemma, the corresponding version of Lemma 1 in the multigraph problem.
\begin{lemma}
\label{cor1.2}
With the same notation and conditions in definition 2, we have for any two $m$-sized sets of samples $\ca Z_1^m$,$\ca Z_1^{m\prime}$ that differ in a single set $\ca Z_i$ drawn from $\ca G$ satisfy:
\begin{equation*}
    \sup_{m\geq 1}\sup_{\ca S^\prime\in\bb S}\sup_{i,j\in\ca V}[ |L(h_{\ca Z_1^m}(\ca T_j^\prime),\bfa Y_j^\prime)-L(h_{\ca Z_1^{ m\prime}}(\ca T_j^\prime),\bfa Y_j^\prime)| ]\leq N\mu,\; \text{ with } S_j^\prime=(\ca T_j^\prime,\bfa Y_j^\prime)\in\ca S^\prime
\end{equation*}
\end{lemma}
Then we start the proof of theorem 2.
 We define $\Phi(\ca Z_1^m)=|\emrr{\ca Z_1^m}-\mrr{\ca Z_1^m}|$. Let $\ca Z_1^{\prime m}$ be another sample drawn from $\ca G$ that differs from $\ca Z_1^m$ by only a single vertex $j_0$ at $\ca Z_{i_0}$ (i.e. $\ca Z_1^m=\ca Z_1^{\prime m}\setminus\{\bfa Z_{i_0}^{(j_0)}\}\cup\{\bfa Z_{i_0}^{\prime (j_0)}\}$. We define set of pairs $\pa S=\{(i,j):i\in\ca V,j\in[m]\}$,  $\pa S_1=\{(i,j):i=i_0,i_0\in\Xi(j)\}$, and $\pa S_2=\pa S\setminus\pa S_1$. Whence we have $ card(\pa S_1)=\ca N_i$ and $card(\pa S_2)=mN-\ca N_i$.

Similar to the proof of theorem 1. The main idea is still Chernoff style concentration inequality. 
\begin{equation*}
\Phi(\ca Z_1^m)-\Phi(\ca Z_1^{\prime m})\leq|\emrr{\ca Z_1^m}-\emrr{\ca Z_1^{\prime m}}|+|\mrr{\ca Z_1^m}-\mrr{\ca Z_1^{\prime m}}|
\end{equation*}
And we bound them seperately, which leads to
\begin{align*}
    \emrr{\ca Z_1^m}-\emrr{\ca Z_1^{\prime m}}&=\frac{1}{mN}\sum_{(i,j)\in\pa S_2}(L(h_{\ca Z_1^m}(\ca T_j^{(i)}),\bfa Y_j^{(i)}) -L(h_{\ca Z_1^{\prime m}}(\ca T_j^{(i)}),\bfa Y_j^{(i)}))\\
    &+\frac{1}{mN}\sum_{(i,j)\in\pa S_1}(L(h_{\ca Z_1^{\prime m}}(\ca T_j^{(i)}),\bfa Y_j^{(i)})-L(h_{\ca Z_1^{\prime m}}(\ca T_j^{\prime (i)}),\bfa Y_j^{^\prime(i)}))\\
    &\leq \frac{(mN-\ca N_i)\mu}{mN}+\frac{\ca N_iB_L}{mN}
\end{align*}
and
\begin{align*}
     \mrr{\ca Z_1^m}-\mrr{\ca Z_1^{\prime m}}\leq\mu
\end{align*}
Then we immediately have:
\begin{equation*}
    \Phi(\ca Z_1^m)-\Phi(\ca Z_1^{\prime m})\leq(2-\frac{d_i}{m})\mu+\frac{\ca N_iB_L}{mN}
\end{equation*}
Using theorem \ref{th1}, we obtained that
\begin{equation}
\label{ineq2}
    \bfa P(\Phi(\ca Z_1^m)-\bb E_{\bb D}[\Phi(\ca Z_1^m)]\geq\epsilon)\leq\exp\bigg(-\frac{(1-\alpha)\epsilon^2}{2m\sum_{i=1}^N((2-\frac{d_i}{m})\mu+\frac{\ca N_iB_L}{mN})^2}\bigg)
\end{equation}

We will bound $\bb E_{\ca Z_1^m\sim\bb D^m}[\Phi(\ca Z_1^m)]$ in what follows. Although we can follow the method in theorem 1, a refinement is obtained through projection. Instead of taking the expectation over $\ca Z_1^m$, we introduced $(\ca Z_1^m, \ca Z^{\text{test}})$ where $\ca Z^{\text{test}} $ is drawn from $\ca G$ according to $\bb D$ that is independent of $\ca Z_1^m$. By the definition of the generalization error:
\begin{equation*}
    \bb E_{\ca Z_1^m\sim\bb D^m}[R(h_{\ca Z_1^m})]=\bb E_{\ca Z_1^m\sim\bb D^m}[\bb E_{\ca Z^{\text{test}}}\wh R_{\ca Z^{\text{test}}}(h_{\ca Z_1^m})]=\bbemt [\wh R_{\ca Z^{\text{test}}}(h_{\ca Z_1^m})]
\end{equation*}
By the linearity of expectation and the property of i.i.d.
\begin{equation*}
    \bbem[\emrr{\ca Z_1^m}]=\frac{1}{m}\sum_{i=1}^m\bbem[\wh R_{\ca Z_i}(h_{\ca Z_1^m})]=\bb E_{\ca Z_1^m\sim\bb D^m}[\wh R_{\ca Z_1}(h_{\ca Z_1^m})]
\end{equation*}
where the second equality comes from permutation. We further have that:
\begin{align*}
    \bb E_{\ca Z_1^m\sim\bb D^m}[\wh R_{\ca Z_1}(h_{\ca Z_1^m})]&=\bbemt[\wh R_{\ca Z^{\text{test}}}(h_{\ca Z_1^{m\prime}})]
\end{align*} with $\ca Z_1^{ m\prime}$ being the  $m$ sets of samples containing $\ca Z^{\text{test}}$ extracted from the  $m+1$ set of samples formed by $\ca Z_1^{m}$ and $\ca Z^{\text{test}}$. 
Let $\ca S^\tet$ be induced by $\ca Z^\tet$ and $S_i^\tet=(\ca T_i^\tet,\bfa Y_i^\tet)\in\ca S^\tet$, we immediately have that
\begin{align*}
    \bbem \Phi(\ca Z_1^m)&=\bbemt [\wh R_{\ca Z^{\text{test}}}(h_{\ca Z_1^m})]-\bbemt[\wh R_{\ca Z^{\text{test}}}(h_{\ca Z_1^{m\prime}})]\\
    &\leq\bbemt[\wh R_{\ca Z^{\text{test}}}(h_{\ca Z_1^m})-\wh R_{\ca Z^{\text{test}}}(h_{\ca Z_1^{m\prime}})]\\
    &=\bbemt[\frac{1}{N}\sum_{j\in\ca V}L(h_{\ca Z_1^m}(\ca T^\tet_j),\bfa Y^\tet_j)- L(h_{\ca Z_1^{m \prime}}(\ca T^\tet_j),\bfa Y^\tet_j)]\\
    &\leq N\mu
\end{align*}
where the last inequality comes from lemma ~\ref{cor1.2} by substituting $\Lambda$ with $\ca V$.

Then we replace $\bbem \Phi(\ca Z_1^m)$ in \ref{ineq2} and complete the proof.
\subsection{Proof of Lemma 5}
In the $\gamma$-strongly convex regime, we can bound the difference of $\bfa w$ with the following lemma.
\begin{lemma}
\label{lm:5}
When $f(\tz{i},\bfa w)$ is $\lambda$-smooth and $\gamma$-strongly convex, we have
\begin{equation*}
    \wt{t}\leq
    \begin{cases}
    \alpha_t^2\lambda\Vert\delta^i\bfa w_{t-1}\Vert+\alpha_tB_Z\zeta &\text{ if $i\in \fk S_{n_t}\text{ and } n_t\neq i\text{ and }\frac{2\alpha_t\lambda\gamma}{\lambda+\gamma}\in[0,1-\alpha_t^4\lambda^2]$}\\
    (1-\frac{\alpha\lambda\gamma}{\lambda+\gamma})\wt{t-1}+\alpha_tB_Z\zeta& \text{ if $i\in \fk S_{n_t}\text{ and } n_t\neq i\text{ and }\frac{2\alpha_t\lambda\gamma}{\lambda+\gamma}\in[1-\alpha_t^4\lambda^2,1]$}\\
    2\alpha_t\ca L+\Vert\delta^i\bfa w_{t-1}\Vert&\text{ if $n_t=i$}\\
    (1-\frac{\alpha\lambda\gamma}{\lambda+\gamma})\Vert\delta^i\bfa w_{t-1}\Vert&\text{ if $i\notin \fk S_{n_t}$}
    \end{cases}
    \end{equation*}
    \end{lemma}
    \begin{proof}
When $i\notin \fk S_{n_t}$, we have
\begin{align*}
    \delta^i\bfa w_t&=\delta^iG(\bfa w,\alpha_t,n_t)=\delta^i\bfa w_{t-1} +\alpha_t\delta^i\nabla f(\tz{n_t},\bfa w_{t-1})
\end{align*}
Hence:\begin{align*}
    \Vert\delta^i\bfa w_t\Vert^2\leq\delta^{i}\bfa w_{t-1}^2+\alpha_t^2\Vert\delta^i\nabla f(\tz{n_t},\bfa w_{t-1})\Vert^2-\alpha_t\langle\delta^i\bfa w_t,\delta^i\nabla f(\tz{n_t},\bfa w_{t-1})\rangle
\end{align*}
Then we use the fact that $f(\bfa w)-\frac{\gamma^2}{2}\Vert\bfa w\Vert^2$ is convex and $\lambda-\gamma$-smooth and by lemma \ref{lm:co} to obtain the following:
\begin{align*}
    &\langle\delta^i\bfa w_t,\delta^i\nabla f(\tz{n_t},\bfa w_{t-1})\rangle\geq(\frac{\lambda\gamma}{\lambda+\gamma})\Vert\delta^i\bfa w_t\Vert^2-\zeta B\Vert\delta^i\bfa w_t\Vert\\
    &+\frac{1}{\lambda+\gamma}\Vert\nabla_{\bfa w} f(\fk S_{n_t}^i(\bfa Z),\bfa w_{t-1})-\nabla_{\bfa w} f(\fk S_{n_t}^i(\bfa Z),\bfa w^i_{t-1})\Vert^2
\end{align*}
We note that:
\begin{align*}
    \delta^i\nabla f(\tz{n_t},\bfa w_{t-1})&=\big[\nabla_{\bfa w} f(\fk S_{n_t}(\bfa Z),\bfa w_{t-1})- \nabla_{\bfa w} f(\fk S_{n_t}^i(\bfa Z),\bfa w_{t-1})+\nabla_{\bfa w} f(\fk S_{n_t}^i(\bfa Z),\bfa w_{t-1})\\
    &-\nabla_{\bfa w} f(\fk S_{n_t}^i(\bfa Z),\bfa w^i_{t-1})\big ]
\end{align*}
and by definition we have
 \begin{align*}
     &\Vert\delta^i\nabla f(\tz{n_t},\bfa w_{t-1})\Vert^2\leq\Vert\nabla_{\bfa w} f(\fk S_{n_t}^i(\bfa Z),\bfa w_{t-1})-\nabla_{\bfa w} f(\fk S_{n_t}^i(\bfa Z),\bfa w^i_{t-1})\Vert^2+B^2\zeta^2\\
     &+2B\zeta\Vert\nabla_{\bfa w} f(\fk S_{n_t}^i(\bfa Z),\bfa w_{t-1})-\nabla_{\bfa w} f(\fk S_{n_t}^i(\bfa Z),\bfa w^i_{t-1})\Vert
 \end{align*}
that leads to
\begin{align*}
    \Vert\delta^i \bfa w_t\Vert_{2}^2&\leq (1-\frac{2\alpha_t\lambda\gamma}{\lambda+\gamma})\Vert\delta^i\bfa w_{t-1}\Vert^2_2-\alpha_t(\frac{2}{\gamma+\lambda}-\alpha_t)\Vert\nabla_{\bfa w} f(\fk S_{n_t}^i(\bfa Z),\bfa w_{t-1})-\nabla_{\bfa w} f(\fk S_{n_t}^i(\bfa Z),\bfa w^i_{t-1})\Vert\\
    &-\zeta B\Vert\delta^i\bfa w_t\Vert+2B\zeta\alpha_t^2\Vert\nabla_{\bfa w} f(\fk S_{n_t}^i(\bfa Z),\bfa w_{t-1})-\nabla_{\bfa w} f(\fk S_{n_t}^i(\bfa Z),\bfa w^i_{t-1})\Vert+\alpha^2_tB^2\zeta^2\\
    &\leq\bigg(1-2\frac{\alpha_t\lambda\gamma}{\gamma+\lambda}\bigg)\Vert\delta^i \bfa w_{t-1}\Vert^2+2B\zeta\alpha_t^2\lambda\Vert\delta^i\bfa w_{t-1}\Vert+\alpha_t^2B^2\zeta^2\\
\end{align*}
Henceforth, when
$
    \alpha_t^4\lambda^2+\frac{2\alpha_t\lambda\gamma}{\lambda+\gamma}\leq 1
$
we have:
\begin{equation*}
     \Vert\delta^i \bfa w_t\Vert_{2}\leq\alpha^2_t\lambda\Vert\delta^i \bfa w_{t-1}\Vert_{2} +\alpha_tB\zeta
\end{equation*}
otherwise when 
$
    \alpha_t^4\lambda^2+\frac{2\alpha_t\lambda\gamma}{\lambda+\gamma}> 1
$ and $\alpha_t\leq\frac{\lambda+\gamma}{\lambda\gamma}$ we have
\begin{equation*}
    \Vert\delta^i \bfa w_t\Vert_{2}\leq\bigg(1-2\frac{\alpha\lambda\gamma}{\lambda+\gamma}\bigg)^{\frac{1}{2}}\Vert\delta^i\bfa w_{t-1}\Vert+\alpha_t B\zeta\leq (1-\frac{\alpha\lambda\gamma}{\lambda+\gamma})\Vert\delta^i\bfa w_{t-1}\Vert+\alpha_tB\zeta
\end{equation*}
\end{proof}
The proof of lemma 5 follows directly from lemma \ref{lm:5} inductively, which adopts the method of ~\citet{hardt2016train}. 
First, we note that:
$
    \bfa P(i\in \fk S_{n_t}\text{ and } n_t\neq i)=\frac{\ca N_i-1}{N} 
$, $
    \bfa P(i=n_t)=\frac{1}{N} 
$ and $
    \bfa P(i\notin \fk S_{n_t})=\frac{N-\ca N_i}{N} 
$ 
Also we denote $\Gamma(j)$ as the time we encounter $\Xi(j)$ s.t. $i\in\Xi(j)\text{ and } i\neq j$ we immediately have
\begin{align*}
    \bfa P(\Gamma(j)=t\text{ and } j\neq i)&=\big(\frac{ N-\ca N_i}{N}\big)^{t-1}\cdot\frac{\ca N_i-1}{N}\\
    \bfa P(\Gamma(i)=t)&=\big (\frac{N-\ca N_i}{N}\big)^{t-1}\cdot\frac{1}{N}\\
    \bfa P(\Gamma(j)< t)&\leq \frac{\ca N_i}{N}\sum_{i=1}^{t-2}\big(\frac{N-\ca N_i}{N}\big)^i=1-\big((1-d_i)\big)^{t-1}
\end{align*}
Hence we have:
\begin{align*}
    \bb E[\Vert\delta^i\bfa w_t\Vert]&\leq\bfa P(\Gamma(j) = t\text{ and } j\neq i)\cdot \alpha_tB\zeta +\bfa P(\Gamma(i)=t)\cdot2\alpha_t\ca L \\
    &+\bfa P(\Gamma(i)<t) \cdot\bigg((d_i-1)\big((1-\alpha^2_t\lambda)\bb E[\Vert\delta^i\bfa w_{t-1}\Vert]+\alpha_tB\zeta\big)\\
    &+(1-d_i)\big(1-\frac{\alpha_t\lambda\gamma}{\lambda+\gamma}\big)\bb E[\Vert\delta^i\bfa w_{t-1}\Vert]+\frac{1}{N}\big(\bb E[\Vert\delta^i\bfa w_{t-1}\Vert]_2+2\alpha_t\ca L\big)\bigg)\\
    &+\bfa P(\Gamma(j)>i)\cdot 0\\
    &\leq\underbrace{\bigg[1-(1-d_i)^{t-1}\bigg]}_{\bfa P_{t,i}}\underbrace{\big(d_i\alpha_t\lambda(\frac{\gamma}{\lambda+\gamma}-\alpha_t)+\frac{\alpha_t^2\lambda}{N}+(1-\frac{\alpha_t\lambda\gamma}{\lambda+\gamma})\big)}_{\pa Z_{t,i}}\bb E[\Vert\delta^i\bfa w_{t-1}\bfa P_{t,i}\Vert]\\
    &+\underbrace{\alpha_t B\zeta(d_i-\frac{1}{N})+2\alpha_t\ca L\frac{1}{N}}_{\pa Y_{t,i}}\\
    &\leq\bfa P_{t,i}\pa Z_{t,i}\bb E[\wt{t-1}] +\pa Y_{t,i}
\end{align*}
Solving the above inequality is equivalent to solving the following:
\begin{equation*}
    \bb E[\Vert\delta^i\bfa w_t\Vert]=\bfa P_{t,i}\pa Z_{t,i}\bb E[\Vert\delta^i\bfa w_t\Vert]+\pa Y_{t,i}
\end{equation*}
In particular, if $\alpha_t=\alpha$ for all $t$, $\pa Z_{t,i}=\pa Z_i$ uniformly for all $t$. We can then obtain the final form of solution through algebraic manipulation.
\subsection{Proof of Theorem 6}
In the proof of high probability bound, we first upper bound the variance of  $\dw$, followed by upper-bounding the supremum by the sum of the random variable. Then, using Chebyshev's inequality, we obtain the high probability upper bound. We first note that:
\begin{align*}
    Var[\Vert\delta^i\bfa w_t\Vert]&=\bb E[\Vert\delta^i\bfa w_t\Vert^2]-\bb E[\Vert\delta^i\bfa w_t\Vert^2]\\
    &\leq\bfa P(\Gamma(j) = t\text{ and } j\neq i)\cdot (\alpha_tB\zeta)^2+\bfa P(\Gamma(i)=t)\cdot(2\alpha_t\ca L)^2 \\
    &+\bfa P(\Gamma(i)<t) \cdot\bb E\bigg [\bigg(\frac{\ca N_i-1}{N}\big((1-\alpha^2_t\lambda)[\Vert\delta^i\bfa w_{t-1}\Vert]+\alpha_tB\zeta\big)\\
    &+\frac{N-\ca N_i}{N}\big(1-\frac{\alpha_t\lambda\gamma}{\lambda+\gamma}\big)[\Vert\delta^i\bfa w_{t-1}\Vert]+\frac{1}{N}\big([\Vert\delta^i\bfa w_{t-1}\Vert]+2\alpha_t\ca L\big)\bigg)^2\bigg ]\\
    &+\bfa P(\Gamma(j)>i)\cdot 0-\bb E[\Vert\delta^i\bfa w_{t}\Vert]^2\\
    &\leq\underbrace{\bigg[1-\bigg(\frac{N-\ca N_i}{N}\bigg)^{t-1}\bigg]}_{\bfa P_0}\bigg(\underbrace{\big(d_i\alpha_t\lambda(\frac{\gamma}{\lambda+\gamma}-\alpha_t)+\frac{\alpha_t^2\lambda}{N}+(1-\frac{\alpha_t\lambda\gamma}{\lambda+\gamma})\big)^2}_{\pa Z^2_{t,i}}\bb E[\Vert\delta^i\bfa w_{t-1}\Vert^2]\\
    &+ 2\underbrace{(\alpha_t B\zeta\frac{\ca N_i-1}{N}+2\alpha_t\ca L\frac{1}{N})}_{\pa Y_{t,i}}\underbrace{\big(d_i\alpha_t\lambda(\frac{\gamma}{\lambda+\gamma}-\alpha_t)+\frac{\alpha_t^2\lambda}{N}+(1-\frac{\alpha_t\lambda\gamma}{\lambda+\gamma})\big)}_{\pa Z_{t,i}}\bb E[\Vert\delta^i\bfa w_{t-1}\Vert]\bigg)\\
    &+\underbrace{(\alpha_t B\zeta\frac{\ca N_i-1}{N}+2\alpha_t\ca L\frac{1}{N})^2}_{\pa Y^2_{t,i}}-\bb E[\Vert\delta^i\bfa w_t\Vert]^2
\end{align*}
For simpler algebra, we take the trivial lower bound that $\inf \bb E[\Vert\delta^i\bfa w_t\Vert]=0$ for all $t\in[T]$. For a similar reason, we take fixed step size $\alpha_t=\alpha$. This gives us:
\begin{align*}
     \vwt{t}\leq \bb E[\dw^2]&\leq \pa Z_{i}^2\vwt{t-1}+2\pa Y_{i}\pa Z_{t,i}\ewt{t-1}{}+\pa Y_{i}^2\\
     &\leq \pa Z_{i}^2\vwt{t-1}+\pa Y_{i}^2\big((\pa Z_i^T-1)\frac{2\pa Z_{i}}{\pa Z_i-1}+1\big)
\end{align*}
To solve the above problem, we further use the following notations.
\begin{equation*}
    \bfa A = \pa Z_{i}^2\;\;\;\bfa B=\pa Z_{i}\;\;\;\bfa C=\frac{2\pa Z_{i}\pa Y_{i}^2}{\pa Z_{i}-1}\;\;\;\bfa D=\pa Y_{i}^2\frac{\pa Z_{i}+1}{1-\pa Z_{i}}
\end{equation*}
 Through some algebraic manipulation, the solution can be obtained as: 
\begin{align*}
    \bb E[\wt{T}^2]&\leq\frac{\bfa C}{\bfa B-\bfa A}(\bfa B^T-\bfa A^T)+\frac{\bfa D}{1-\bfa A}(1-\bfa A^T)\\
    &\leq (2\pa Y_{i}^2)\frac{\pa Z_{i}^{2T}-\pa Z_{i}^T}{\pa Z_{i}^2-\pa Z_{i}}+\pa Y_{i}^2\frac{1-\pa Z_{i}^T}{(1-\pa Z_{i})^2}
\end{align*}
Hence, using the sum to upper bound the supremum, we have
\begin{equation*}
    Var[\sup\dw]\leq\bb E[\sup_{i\in\ca V}\dw^2]\leq\sum_{i=1}^N\bb E[\dw^2]=\sum_{i=1}^N\bigg((2\pa Y_{i}^2)\frac{\pa Z_{i}^{2T}-\pa Z_{i}^T}{\pa Z_{i}^2-\pa Z_{i}}+\pa Y_{i}^2\frac{1-\pa Z_{i}^T}{(1-\pa Z_{i})^2}\bigg)
\end{equation*}
Using Chebyshev's inequality, given $\epsilon\geq\sqrt{\sum_{i=1}^N\bigg((2\pa Y_{i}^2)\frac{\pa Z_{i}^{2T}-\pa Z_{i}^T}{\pa Z_{i}^2-\pa Z_{i}}+\pa Y_{i}^2\frac{1-\pa Z_{i}^T}{(1-\pa Z_{i})^2}\bigg)}$, we immediately have 
\begin{equation*}
    \bfa P(\sup_i\Vert\delta^i\bfa w_T\Vert-\bb E\sup_i\Vert\delta^i\bfa w_T\Vert\geq \epsilon)\leq\frac{Var[\sup_i\Vert \delta^i\bfa w_T\Vert]}{\epsilon^2}
\end{equation*}
which holds when $\epsilon\geq\sup_i\sqrt{Var[\Vert\delta^i\bfa w_t\Vert]}$.
Or, with probability at least $1-\delta$, the following holds:
\begin{align*}
    0\leq \sup_i\Vert\delta^i\bfa w_T\Vert&\leq\bb E\sup_i\Vert\delta^i\bfa w_T\Vert+\sqrt{\frac{Var[\sup_i\Vert\delta^i\bfa w_T\Vert]}{\delta}}\\
    &\leq \sup_i\bigg[(\pa Z_i^T-1)\frac{\pa Y_i}{\pa Z_i-1}\bigg]+\sqrt{\frac{1}{\delta}\sum_{i=1}^N\bigg( (2\pa Y_{i}^2)\frac{\pa Z_{i}^{2T}-\pa Z_{i}^T}{\pa Z_{i}^2-\pa Z_{i}}+\pa Y_{i}^2\frac{1-\pa Z_{i}^T}{(1-\pa Z_{i})^2}\bigg)}
\end{align*}
Note that by the conditions on $\gamma$-smoothness and $\lambda$-strongly convexity:
\begin{equation*}
  \langle\nabla f(\fk S_j,\bfa w_t),\delta^i\bfa w_T\rangle+\frac{\gamma}{2}\Vert\delta^i\bfa w_T\Vert^2\leq |f(\tz{j},\bfa w^i_T)-f(\tz{j},\bfa w_T)|\leq\langle\nabla f(\fk S_j,\bfa w_t),\delta^i\bfa w_T\rangle+\frac{\lambda}{2}\Vert\delta^i\bfa w_T\Vert^2
\end{equation*}
which indicates that with fixed $\sup_i\dwt$, we have $\beta_2$ is subgaussian with $\sigma = \frac{1}{4}(\lambda-\gamma)^2\dwt^4$, which implies that
\begin{equation*}
    \bfa P(\beta_2-\bb E[\beta_2]\geq\epsilon)\leq\exp\bigg (\frac{-8\epsilon^2}{(\lambda-\gamma)^2(\sup_i\dwt)^4}\bigg )
\end{equation*}
Hence, with probability at least $1-\delta$, given $\sup_i\dwt$, we have the following
\begin{equation*}
    \beta_2\leq\bb E[\beta_2]+(\lambda-\gamma)(\sup_i\dwt)^2\sqrt{\frac{\log\frac{1}{\delta}}{8}}
\end{equation*}
Together with union bound we complete the proof.

\subsection{Proof of Lemma 7}
Without the convexity, the algorithm does not have guaranteed convergence. However, the generalization guarantees of the hypothesis only depends on the multi-fidelity stability of the algorithm from which it is returned, regardless of the convergence of the algorithm.
\begin{lemma}
When $f(\tz{i},\bfa w)$ is $\lambda$-smooth and non-convex, we have
\begin{equation*}
    \wt{t}\leq
    \begin{cases}
    (1+\alpha_t\lambda)\wt{t-1}+\alpha_tB_Z\zeta& \text{ if $i\in \fk S_{n_t}$ and  $n_t\neq i$}\\
    2\alpha_t\ca L+\wt{t-1}&\text{ if $n_t=i$}\\
    (1+\alpha_t\lambda)\Vert\delta^i\bfa w_{t-1}\Vert&\text{ if $i\notin \fk S_{n_t}$}
\end{cases}
\end{equation*}
\end{lemma}
The proof follows a similar fashion to the convex case with minor modification.
\begin{align*}
    \bb E[\wt t]&\leq\bfa P(\Gamma(j)=t\text{ and } j\neq i)\cdot \alpha_tB\zeta+\bfa P(\Gamma(i)=t)\cdot 2\alpha_t\ca L\\
    &+\bfa P(\Gamma(i)\leq t)\cdot\bigg(\frac{\ca N_i-1}{N}\big((1+\alpha_t\lambda)\bb E[\wt{t-1}]+\alpha_tB\zeta\big)\\
    &+\frac{N-\ca N_i}{N}(1+\alpha_t\lambda)\bb E[\wt{t-1}]+\frac{1}{N}(\bb E[\wt{t-1}]+2\alpha_t\ca L)\bigg)\\
    &\leq\underbrace{\bigg[1-\bigg(\frac{N-\ca N_i}{N}\bigg)^{t-1}\bigg]}_{\bfa P_0}\underbrace{\frac{N-1}{N}\alpha_t\lambda}_{\pa M_{t}}\bb E[\wt{t-1}]+\underbrace{\alpha_tB\zeta\frac{\ca N_i-1}{N}+2\alpha_t\ca L\frac{1}{N}}_{\pa Y_{t,i}}\\
    &\leq\pa M_t\bb E[\wt{t-1}]+\pa Y_{t,i}
\end{align*}
We then set a fixed step size $\alpha_t=\alpha$ and solve the corresponding difference equation to complete the proof. Note that when $\pa M=0$, the regime collapsed to the sum of arithmetic sequence with interval, $\ca L\pa Y_i$. 

\subsection{Proof of Theorem 8}
The proof also goes in a similar fashion to the convex case. We have for the second order moment
\begin{align*}
    \bb E[\Vert\delta^i\bfa w_t\Vert^2]&\leq\bfa P(\Gamma(j)=t\text{ and } j\neq i)\cdot(\alpha_tB\zeta)^2+\bfa P(\Gamma(i)=t)\cdot(2\alpha_t\ca L)^2\\
    &+\bfa P(\Gamma(i)<t)\cdot \bb E\bigg [\bigg(\frac{\ca N_i-1}{N}\big((1+\alpha_t\lambda)\wt{t-1}+\alpha_tB\zeta\big)+(1-d_i)(a+\alpha_t\lambda)\bigg ]\\
    &\leq \underbrace{(1-\frac{1}{N})^2\alpha^2_t\lambda^2}_{\pa M_{t}^2}\bb E[\wt{t-1}^2]+2\pa Y_{t,i}\underbrace{(1-\frac{1}{N})\alpha_t\lambda}_{\pa M_{t}}\bb E[\wt{t-1}]+\pa Y_{t,i}^2
\end{align*}
Given that step size fixed as $\alpha$ we immediately obtain the following
\begin{align*}
    \bb E[\wt{t-1}^2]&\leq \pa M^2\bb E[\wt{t-1}^2]+2\pa Y_{i}\pa M\bb E[\wt{t-1}]+\pa Y_{i}^2
\end{align*}
To solve the above inequality, we solved the corresponding difference equality. We denote the following variables
\begin{align*}
     \bfa A=\pa M^2\;\;\;\bfa B=\pa M\;\;\;\bfa C= \frac{2\pa Y^2_{i}}{\pa M-1}\;\;\;\bfa D=\pa Y_{i}^2\frac{1+\pa M}{1-\pa M}
\end{align*}
With some algebraic manipulation, the final solution to the difference inequality yielded
\begin{align*}
    \bb E[\wt{T}^2]&\leq\frac{\bfa C}{\bfa B-\bfa A}(\bfa B^T-\bfa A^T)+\frac{\bfa D}{1-\bfa A}(1-\bfa A^T)\\
    &\leq(2\pa Y_{i}^2)\frac{\pa M^{2T}-\pa M^T}{\pa M^2-\pa M}+\pa Y_{i}^2\frac{1-\pa M^T}{(1-\pa M)^2}
\end{align*}
Hence we immediately have
\begin{equation*}
    \bb E[\sup_i\dw^2]\leq\sum_{i=1}^N\bb E[\sup_i\dw^2]=\sum_{i=1}^N\bigg((2\pa Y_{i}^2)\frac{\pa M^{2T}-\pa M^T}{\pa M^2-\pa M}+\pa Y_{i}^2\frac{1-\pa M^T}{(1-\pa M)^2}\bigg)
\end{equation*}
By Chebyshev's inequality, we have:
\begin{align*}
    \bfa P(\sup_i\wt{T}-\bb E[\sup_i\wt{T}]\geq\epsilon)\leq\frac{\bb E[\sup_i\dw^2]}{\epsilon^2}\leq\frac{\sum_{i=1}^N\bigg((2\pa Y_{i}^2)\frac{\pa M^{2T}-\pa M^T}{\pa M^2-\pa M}+\pa Y_{i}^2\frac{1-\pa M^T}{(1-\pa M)^2}\bigg)}{\epsilon^2}
\end{align*}
which gives us that with probability at least $1-\delta$ and for $\epsilon\geq\sqrt{\sum_{i=1}^N\bigg((2\pa Y_{i}^2)\frac{\pa M^{2T}-\pa M^T}{\pa M^2-\pa M}+\pa Y_{i}^2\frac{1-\pa M^T}{(1-\pa M)^2}\bigg)} $
\begin{align*}
    \sup_i\wt{T}&\leq(\pa M_{t}^T-1)\frac{\sup_i\pa Y_{i}}{\pa M_{t}-1}+\sqrt{\frac{1}{\delta}\sum_{i=1}^N\bigg(2\pa Y_{i}^2\frac{\pa M_{t}^{2T}-\pa M_{t}^T}{\pa M_{t}^2-\pa M_{t}}+\pa Y_{i}^2\frac{1-\pa M_{t}^T}{(1-\pa M_{t})^2}\bigg)}
\end{align*}
Given
\begin{align*}
    0\leq|\beta_2|\leq\ca L\sup_i\wt{T}
\end{align*}
we have that $\beta_2$ is sub-Gaussian with $\sigma^2=\frac{1}{4}\ca L^2(\sup_i\wt{T})^2$. Given $\sup_i\wt{T}$, for $\epsilon\geq 0$:
\begin{equation*}
\bfa P(\beta_2-\bb E[\beta_2]\geq\epsilon)\leq\exp\bigg(\frac{-2\epsilon^2}{\ca L^2(\sup_i\wt{T})^2}\bigg)
\end{equation*}
The above is analogous to that with probability at least $1-\delta$ :
\begin{equation*}
   \beta_2\leq\bb E[\beta_2]+(\ca L\sup_i\wt{T})\sqrt{\frac{\log\frac{1}{\delta}}{2}}
\end{equation*}
By union bound and replacing the first moment and we complete the proof.

\subsection{Proof of Theorem 9}
We let $\Pi_{\bb A}\in\bb R^{N\times N}$ be the boolean projection matrix such that $\Pi_{\bb A}\otimes\bfa B\in\bb A$ for all $\bfa B\in\bb R^{N\times N}$. Let $\tda=\Pi_{\bb A}\otimes\wha$ with $\wha\in\bb R^{N\times N}$. Then the first order derivative w.r.t. $\wha$ is given by:
\begin{equation*}
    \frac{\partial f(\tda)}{\partial \wha_{i.j}}=\text{Tr}\bigg[( \frac{\partial f(\tda)}{\partial \tda_{i.j}})^\top  \frac{\partial \tda}{\partial \wha_{i.j}}\bigg]=\frac{\partial f(\tda)}{\partial \tda_{i,j}}\bfa 1_{\tda_{i,j}\neq 0}
\end{equation*}.

The first order condition yields that
\begin{align*}
    \Pi_{\bb A}\otimes[-\bfa y\bfa w^\top\bfa X^\top+\tda(\bfa {Xww^\top X^\top}+\gamma\bfa I)]=\bfa 0
\end{align*}
which gives the following closed form solution
\begin{align}\label{aa}
    \tda &=\Pi_{\bb A}\otimes(\bfa{yw^\top X^\top}(\bfa X\bfa w\bfa w^\top\bfa X^\top+\gamma\bfa I)^{-1})
\end{align}
hence
\begin{equation}
\label{up}
    \Vert\tda\Vert=O(\sup_i d_i)
\end{equation}
Then we give a lower bound on the $i$-th stability. 
\begin{align*}
    \beta^i&=\sup_{j,\bfa Z,\bfa Z^i}\sup_{\ca S=(\bfa X^\prime,\bfa y^\prime)}[(\tda\bfa X^\prime\bfa w)_j-\bfa y^\prime_j)^2-((\tda^i\bfa X^\prime\bfa w)_j-\bfa y^\prime_j)^2]\\
    &= \sup_{j,\ca S,\bfa Z,\bfa Z^i}\big [ (\Delta_i\tda\bfa X^\prime\bfa w)_j((\tda+\tda^i)\bfa X^\prime\bfa w-2\bfa y^\prime_j)_j\big ]\\
    &=\Theta(B_y\sup_{\bfa X^\prime,\bfa Z,\bfa Z^i}\Vert\Delta_i\tda\bfa X^\prime\bfa w\Vert_\infty)\\
    &=\Omega(\sup_{\bfa Z,\bfa Z^i}\Vert\Delta_i\tda\Vert_{2.\infty})
\end{align*}
Moreover, applying \ref{aa} and coupled with the fact that for all $f, \;\sup_{\bfa x,\bfa y} f(\bfa x,\bfa y)\geq\sup_{\bfa y}f(\bfa x,\bfa y)$ we have
\begin{align*}
    \sup_{\bfa Z,\bfa Z^i}\Vert\Delta_i \tda\Vert_{2,\infty} &=\sup_{\bfa Z,\bfa Z^i}\Vert(\tda^i-\tda)\Vert_{2,\infty}\\
    &\geq\sup_{\bfa y,\bfa y^i}\Vert\Pi_{\bb A}(2B_y\bfa 1_{i}\bfa w^\top\bfa X^\top(\bfa X\bfa w\bfa w^\top\bfa X^\top+\gamma\bfa I)^{-1} )\Vert_{2,\infty}\\
    &=\Omega(d_i)
\end{align*}
Hence
\begin{equation}
\label{lo}
    \beta_2=\sup_i\beta^i=\Theta(\sup_i d_i)
\end{equation}
Given the upper bound at \ref{up} and lower bound at \ref{lo}, we complete the proof.

\subsection{Proof of Theorem 10}
We consider two special cases with simple algebra and leave the general case as an open problem.
\begin{compactenum}
    \item $\bfa y^i=\bfa y$, $\bfa X^i\neq\bfa X$ and $\Vert\bfa X^i-\bfa X\Vert\ll\Vert\bfa X\Vert$, which we coined first order $\bfa X$ perturbation.
    \item $\bfa y^i\neq\bfa y$ and $\bfa X^i=\bfa X$, which we coined $\bfa y$ perturbation.
\end{compactenum}
To lowerbound the discrepancy, we will need the following observation
\begin{equation*}
     \beta_2-\beta_1\geq\inf_i(\beta^i-\alpha^i)
\end{equation*}
To bound the r.h.s,
Assuming that 
\begin{equation*}
    (\bfa Z^*,\bfa Z^{*i},\bfa X^*,\bfa y^*)=\argmax_{\bfa Z,\bfa Z^{i},\bfa X^\prime,\bfa y^\prime}\sup_{j\notin\Xi(i)}|(\tda\bfa X^\prime\bfa w)_j-\bfa y^\prime_j)^2-((\tda^i\bfa X^\prime\bfa w)_j-\bfa y^\prime_j)^2|
\end{equation*} and let $\tda^*,\tda^{*i}$ be the returned hypothesis of the optimization problem on $\bfa Z$ and $\bfa Z^i$. 
 Given those notations, we lower bound the gap between $\beta^i$ and $\alpha^i$ by
\begin{align*}
    \beta^i-\alpha^i&\geq \sup_{j}|(\tda^*\bfa X^{*\prime}\bfa w)_j-\bfa y^{*\prime}_j)^2-((\tda^{*i}\bfa X^{*\prime}\bfa w)_j-\bfa y^{*\prime}_j)^2|\\
    &-\sup_{\pr j\notin\Xi(i)}|(\st\tda\stp{\bfa X}\bfa w)_{\pr j}-\stp{\bfa y}_{\pr j})^2-(\tda^{*i}\stp{\bfa X}\bfa w)_{\pr j}- \stp{\bfa y}_{\pr j})^2|\\
    &=\sup_{j}\bigg|\bigg((\Delta_i\tdas\stp{\bfa X}\bfa w)\otimes((\st\tda+\stp\tda)\stp{\bfa X}\bfa w-2\stpb y)\bigg)_j\bigg|\\
    &-\sup_{j^\prime\notin\Xi(i)}\bigg|\bigg((\Delta_i\tdas\stp{\bfa X}\bfa w)\otimes((\st\tda+\stp\tda)\stp{\bfa X}\bfa w-2\stpb y)\bigg)_{j^\prime}\bigg|
\end{align*}

\textbf{First order perturbation in $\bfa X$}

Given \ref{aa}, we have
\begin{align*}
    \Delta_i\tda^*\approx\Pi_{\bb A}\otimes\bfa y\bfa w^\top\Delta_i\bfa X^{*\top}(\bfa X\bfa w\bfa w^\top\bfa X^\top+\gamma\bfa I)^{-1}
\end{align*}
hence $ \Delta_i\tda^*_{i^\prime,j^\prime}\neq 0$ only if $i=i^\prime$ and $j\in\Xi(i)$. Whence $(\Delta_i\tdas\stp{\bfa X}\bfa w)_{j}\neq 0$ only if $j\in\Xi(i)$, which indicates that
\begin{align*}
\beta^i-\alpha^i&\geq\sup_{j}\bigg|\bigg((\Delta_i\tdas\stp{\bfa X}\bfa w)\otimes((\st\tda+\stp\tda)\stp{\bfa X}\bfa w-2\stpb y)\bigg)_j\bigg|\\
    &-\sup_{j^\prime\notin\Xi(i)}\bigg|\bigg((\Delta_i\tdas\stp{\bfa X}\bfa w)\otimes((\st\tda+\stp\tda)\stp{\bfa X}\bfa w-2\stpb y)\bigg)_{j^\prime}\bigg|\\
    &=\sup_{j\in\Xi(i)}\bigg|\bigg((\Delta_i\tdas\stp{\bfa X}\bfa w)\otimes((\st\tda+\stp\tda)\stp{\bfa X}\bfa w-2\stpb y)\bigg)_j\bigg|-0\\
    &=\Omega(d_i)
\end{align*}
Hence $\beta_2-\beta_1=\Omega(\sup_i d_i)$

\textbf{Perturbation on $\bfa y$}

Given \ref{aa},
 $\Delta_i\tdas$ can be written in the following form:
\begin{equation*}
    \Delta_i\tdas=\Pi_{\bb A}(2|\Delta\stpb y|\bfa 1_i\bfa w^\top\bfa X^\top(\bfa X\bfa w\bfa w^\top\bfa X^\top+\gamma\bfa I)^{-1} )
\end{equation*} which will only have nonzero value at $i$-th row.
Then we immediately have:
\begin{align*}
     &\beta_i-\alpha_i\geq\sup_{j}\bigg|\bigg((\Delta_i\tdas\stp{\bfa X}\bfa w)\otimes((\st\tda+\stp\tda)\stp{\bfa X}\bfa w-2\stpb y)\bigg)_j\bigg|\\
    &-\sup_{j^\prime\notin\Xi(i)}\bigg|\bigg((\Delta_i\tdas\stp{\bfa X}\bfa w)\otimes((\st\tda+\stp\tda)\stp{\bfa X}\bfa w-2\stpb y)\bigg)_{j^\prime}\bigg|\\
    &=\bigg((\Delta_i\tdas\stp{\bfa X}\bfa w)\otimes((\st\tda+\stp\tda)\stp{\bfa X}\bfa w-2\stpb y)\bigg)_i\\
    &=\Omega(\frac{1}{N})
\end{align*}
which completes the proof.
\end{document}